\documentclass[conference]{IEEEtran}
\IEEEoverridecommandlockouts
\usepackage{cite}
\usepackage{amsmath,amssymb,amsfonts,amsthm}
\usepackage{algorithmic}
\usepackage{graphicx}
\usepackage{textcomp}
\usepackage{xcolor}
\usepackage{booktabs}
\def\BibTeX{{\rm B\kern-.05em{\sc i\kern-.025em b}\kern-.08em
    T\kern-.1667em\lower.7ex\hbox{E}\kern-.125emX}}

\newcommand{\R}{\mathbb{R}}
\newcommand{\C}{\mathbb{C}}
\newcommand{\Z}{\mathbb{Z}}
\newcommand{\AHI}{\text{AHI}}

\newtheorem{definition}{Definition}[section]

\begin{document}

\title{Deep Learning Approaches for Sleep Apnea Classification from
	Polysomnographic EEG Signals}

\author{
	\IEEEauthorblockN{Shashank Manjunath and Mukesh Cheemakurthi}
	\IEEEauthorblockA{\textit{Khoury College of Computer Sciences} \\
		\textit{Northeastern University} \\
		Boston, USA \\
		\texttt{\{manjunath.sh, cheemakurthi.m\}@northeastern.edu}}

	% \and
	%
	% \IEEEauthorblockN{Mukesh Cheemakurthi}
	% \IEEEauthorblockA{\textit{Khoury College of Computer Sciences} \\
	%   \textit{Northeastern University}\\
	%   Boston, USA \\
	%   % \texttt{cheemakurthi.m@northeastern.edu}
	% }

	\and

	\IEEEauthorblockN{Aarti Sathyanarayana}
	\IEEEauthorblockA{\textit{Khoury College of Computer Sciences} \\
		\textit{Bouv\'e College of Health Sciences}\\ Northeastern University \\
		Boston, USA \\
		\texttt{a.sathyanarayana@northeastern.edu}}
}

\maketitle

\begin{abstract}
	Sleep apnea diagnosis via polysomnography remains resource intensive and
	relies on time consuming manual data analysis and scoring. Recent work has
	demonstrated that central nervous system effects of sleep apnea events can be
	detected through electroencephalogram (EEG) signals. However, most work uses
	a single feature type on various datasets combined with different
	classification algorithms. In this work, we present a comprehensive
	comparison of deep learning architectures and feature representations for
	automated sleep apnea detection from multichannel EEG on a single dataset of
	pediatric subjects. We evaluate Vision Transformers and Graph Attention
	Networks across distinct signal representations: raw temporal signals,
	short-time Fourier transform spectrograms, coherence based graphs, and two
	topological data analysis (TDA) derived features. Using age and sex matching
	of our train and test sets, we train on 2410 pediatric subjects and test on
	575 pediatric subjects. We achieve a best test AUC of 0.750 using a vision
	transformer based model trained on TDA features. Stratified analysis across
	patient demographics (age, sex, AHI severity) and sleep stages (N1, N2, N3,
	REM) reveals significant performance variation. Our results demonstrate the
	feasibility of EEG based automated OSA screening while highlighting essential
	challenges for clinical deployment.
\end{abstract}

\section{Introduction}

Obstructive sleep apnea (OSA) is a highly prevalent sleep disorder affecting an
estimated 936 million adults and up to 6.4\% children
worldwide~\cite{benjafieldEstimationGlobalPrevalence2019,
	magnusdottirPrevalenceObstructiveSleep2024}. Sleep apnea is characterized by
repeated episodes of complete or partial upper airway obstruction during sleep.
OSA leads to intermittent hypoxia, sleep fragmentation, and excessive daytime
sleepiness. Left untreated, OSA significantly increases the risk of
cardiovascular disease and all-cause
mortality~\cite{levyObstructiveSleepApnoea2015}.

Despite its clinical significance, OSA remains
underdiagnosed~\cite{gouthroPediatricObstructiveSleep2025}. Current diagnosis
relies on overnight polysomnography (PSG), which monitors multiple physiological
signals including electroencephalogram (EEG), electrocardiogram (ECG),
respiratory effort, and oxygen saturation. While PSG studies are the gold
standard for OSA classification, these studies requires expensive equipment,
dedicated sleep laboratories, and trained technicians to manually score sleep
stages and respiratory events~\cite{slowikObstructiveSleepApnea2025}. This
process is time consuming, labor intensive, and subject to interrater
variability~\cite{leeInterraterReliabilitySleep2022}. These limitations create
significant bottlenecks in diagnostic workflows and restrict access to timely
OSA diagnosis, particularly in resource-limited settings.

While OSA is fundamentally a respiratory disorder, the symptoms and effects of
sleep apnea manifest through central nervous system (CNS) dysfunction.
Recurrent apnea events trigger brief awakenings that fragment sleep
architecture and prevent restorative deep sleep and REM sleep. These repeated
arousals, coupled with intermittent hypoxia, lead to chronic sleep disruption
that manifests as excessive daytime sleepiness, impaired cognitive function,
mood disturbances, and reduced quality of life~\cite{urbanoLinkPediatricObstructive2021a}.
Moreover, emerging evidence suggests that chronic OSA-induced hypoxia may cause
long-term neurological damage, including gray matter loss in regions critical
for memory and executive function~\cite{maceyBrainMorphologyAssociated2002}.

Electroencephalograms (EEGs) directly measure these CNS effects through
electrical activity of the brain, providing unique insights into the effect of
apnea events on the brain. Unlike peripheral sensors that detect respiratory
obstruction (nasal airflow, thoracic effort) or its immediate physiological
consequences (oxygen desaturation), EEG captures the signatures of sleep
disruption such as arousal patterns, frequency band alterations, and sleep stage
fragmentation.

Recent advances in machine learning and deep learning have demonstrated promise
for automated sleep disorder detection from physiological
signals~\cite{tyagiSystematicReviewAutomated2023}. EEG signals are particularly
informative for OSA detection, as apnea events trigger changes in EEG signal
patterns~\cite{barnesDetectionSleepApnea2022a,manjunathDetectionSleepOxygen2024a}.
However, automated OSA classification based on EEG signals presents technical
challenges. EEG signals from multichannel recordings present high-dimensional,
complex spatiotemporal patterns that require sophisticated feature extraction.
Furthermore, model performance must generalize across heterogeneous patient
populations with varying demographics, comorbidities, and OSA severity levels
to achieve clinical utility.

While several studies have applied deep learning to OSA detection using
single-channel ECG and other physiological signals, fewer have comprehensively
evaluated multichannel EEG approaches across diverse feature
representations~\cite{srivastavaApneaNetHybrid1DCNNLSTM2023}. Recent work has
explored convolutional neural networks for single-channel
EEG~\cite{barnesDetectionSleepApnea2022a}, YOLO architectures for EEG
spectrograms~\cite{tanciClassificationSleepApnea2025}, and traditional machine
learning with hand-crafted features~\cite{zhaoClassificationSleepApnea2021a}.
However, these studies typically focus on a single feature type or architecture,
limiting understanding of which signal representations best capture apnea
signatures in EEG signals. Furthermore, most prior work reports only aggregate
performance metrics without stratifying results by patient demographics or sleep
stages, which is an essential consideration for clinical deployment given known
variations in OSA presentation across age, sex, and disease severity.

This work addresses these gaps through a comprehensive evaluation of deep
learning architectures and feature representations for OSA detection from
multichannel EEG. We investigate four distinct signal representations: (1) raw
temporal signals , (2) short-time Fourier transform (STFT) spectrograms
capturing time-frequency dynamics, (3) coherence-based graph representations
encoding inter-channel functional connectivity, and (4) topological data
analysis (TDA) features derived from persistent homology. TDA has shown promise
in characterizing sleep-related EEG patterns~\cite{sathyanarayanaTopologicalDataAnalysis2025a}.

We evaluate feature representations using two deep learning architectures:
Vision Transformers (ViT) adapted for physiological signals, and Graph Attention
Networks (GAT) for connectivity-based
representations~\cite{demirEEGGATGraphAttention2022a}. To ensure clinical
relevance, we conduct rigorous stratified analysis across patient demographics
(age groups, sex, AHI severity levels) and sleep stages (N1, N2, N3, REM),
identifying critical factors that influence detection performance in real-world
scenarios.

Our main contributions are threefold. First, we present the first comprehensive
comparison of multiple deep learning architectures and feature representations
for multichannel EEG-based OSA detection. Second, we introduce TDA-based
features (HEPC and AP FAPC) to OSA detection and demonstrate their effectiveness
in capturing apnea signatures compared to traditional representations. Third,
we conduct extensive demographic and sleep stage stratification analysis,
revealing performance disparities across subpopulations that have critical
implications for equitable and robust clinical deployment.

\section{Related Work}

There has been significant recent work on sleep apnea detection from EEG
signals. Prior work by Mahmud et al. (2021) found that variational mode
decomposition can successfully decompose EEG signals for sleep apnea
classification from EEG signals alone~\cite{mahmudSleepApneaDetection2021}.
Combined with a CNN-BiLSTM deep learning model, the authors achieved average F1
scores greater than 0.90 across three datasets of adult subjects using a
subject-independent evaluation method. While this work achieved high F1 scores,
the datasets used were small, with the largest dataset containing 25 mostly male subjects.
Other work by Barnes et al. (2022) developed a CNN for
single-channel sleep apnea detection~\cite{barnesDetectionSleepApnea2022a}. This
model achieved an accuracy of 69.9\% and an average AUC across 10 folds of 0.804
on three datasets of adult subjects. Further work by Zhao et al. (2021) used
traditional deep learning models on two EEG channels combined with manual
feature extraction on 30 subjects~\cite{zhaoClassificationSleepApnea2021a}.
While this study achieved good accuracy of 88.99\%, little information about the demographics of the
training and testing set was reported. Work by Tanci et al. (2025) attempted to use an
object detection network, YOLOv8, on Short-time Fourier Transform (STFT) EEG
features to detect sleep apnea events, achieving an overall accuracy of
93.7\%~\cite{tanciClassificationSleepApnea2025}. The subject demographics are
not reported, but the authors stratified their model results based on sleep apnea
severity, revealing consistent F1 scores across sleep apnea severity. Recent work by
Irurueta et al. (2025) achieved a best AUC of 0.689 for EEG-only detection of
sleep apnea events using a CNN on EEG
signals~\cite{iruruetaAutomaticSleepApnea2025}. Other work by Manjunath et al.
(2024) used Short-time Fourier transforms and CNNs to predict oxygen
desaturations from 30-second EEG signals in pediatric patients, achieving a best
balanced accuracy of 66.8\%~\cite{manjunathDetectionSleepOxygen2024a}.
Critically, this study was conducted on pediatric patients rather than adults,
and accounted for age and sex in the training and testing splits.

\section{Methods}

\subsection{Dataset}

This work uses the Nationwide Children's Hospital Sleep DataBank (NCHSDB)
dataset with 3984 PSG studies from 3673 pediatric
patients~\cite{leeNCHSleepDataBank,leeLargeCollectionRealworld2022d,goldbergerPhysioBankPhysioToolkitPhysioNet2000a}.
This dataset contains full PSG studies, including EEG, pulse oximetry, and nasal
and oral sensors, among others. The dataset furthermore contains technician
labels for sleep stage and sleep breathing events, including oxygen
desaturations, and central, mixed, and obstructive apnea events. We use studies
collected from subjects with 256 Hz EEG sampling rate and 7-channel EEG studies.
In this work, we use 30-second groups of 7-channel EEG signals sampled at 256
Hz. After filtering our dataset to subjects with the appropriate 7 EEG
channels and 256 Hz sampling rate, we were left with 2985 subjects. We label a
particular 30-second window as "disordered breathing" (positive) or "normal
breathing" (negative) if the apnea, hypopnea, oxygen desaturation label in the
positive case or normal breathing label in the negative case covers 75\% of the
window. Data was split by subject, i.e. no data from one subject was in the
training and validation sets. Furthermore, we perform an age and sex stratified
train/validation split, meaning that the training set and the validation set
have the same ratio of each age and sex group. Based on this labeling strategy,
our training set contains 71,547 positive and 2,130,666 total labels (3.36\%
positive), while the test set contains 17,066 positive and 505,817 total labels
(3.37\% positive).

Apnea-hypopnea index (AHI) is a common metric used to classify sleep apnea
severity. It is calculated by counting the total number of apnea and hypopnea
events over a night of sleep, then dividing by the total sleep time. In
pediatric subjects, $\AHI < 1$ is considered healthy, $1 \leq \AHI < 5$ is
considered mild sleep apnea, $5 \leq \AHI < 10$ is considered moderate sleep
apnea, and $\AHI \geq 10$ is considered severe sleep
apnea~\cite{saviniAssessmentObstructiveSleep2019a}. We use these criteria to
stratify subjects by sleep apnea severity in
Section~\ref{subsec:severity_results}.

\subsection{Feature Extraction Approaches}

\subsubsection{Raw EEG Signals}

Our first feature representation are raw EEG signals. We filter powerline noise
with a 60 Hz and 120 Hz 3rd order Butterworth bandstop filter. Since we are
focused on 30-second groups of 7-channel EEG data collected at 256 Hz, this
feature yields a feature array $\mathcal{F} \in \R^{7 \times 7681}$. We apply
no further filtering to the raw signals, and do not address baseline drift or
apply any data deletion or interpolation procedures.

\subsubsection{Short-Time Fourier Transform (STFT)} \label{subsec:stft}

The second featurization technique we test are short-time Fourier transforms. We
take our raw EEG signals and perform powerline noise filtration, then calculate
an STFT with 129 frequency bins (0.5 - 128 Hz) using a Hann window function.
Specifically, we calculate the following discrete STFT for each channel $m$:

\begin{equation}
	\mathcal{S}_m [q, p] = \sum\limits_{k=0}^{N-1} X_m[k] \overline{w[k - ph]} e^{-i 2 \pi q k /N}
\end{equation}

\noindent where $q, p \in \Z$ denote the time and frequency index respectively  and
$\overline{w}$ is the complex conjugate of the Hann window function, and $X_m$
is the discretely sampled $m$th channel of the signal. This process yields
$\mathcal{S} \in \C^{129 \times 62}$, a complex-valued array with 129 frequency
bins and 61 time bins. The STFT array magnitude is log-transformed and
Z-normalize yielding an array $\mathcal{F} \in \R^{7 \times 129 \times 62}$
corresponding to 7 channels

\subsubsection{Coherence-Based Graph Construction} \label{subsec:coherence}

To test out features which integrate functional connectivity metrics, we
construct a graph representation based on our multichannel EEG signal. The graph
contains 7 nodes, one for each EEG of the 7 electrodes. The signal coherence
between any two channels represents the weight of the edge between the two
corresponding nodes, and the STFT of that channel, calculated as in
Subsection~\ref{subsec:stft} are the node features. To calculate coherence, we
take our set of time-series signals $X_i (t)$, where $i \in V = \{1, \cdots, 7\}$
represents the channels and $t \in \{1, \cdots, T\}$ are the time samples of our
signal. We first calculate a smoothed periodogram $\hat f_{m, n}$ between
channels $m$ and $n$. We then calculate the magnitude squared signal coherence
between all channels $m, n$ as follows:

\begin{equation}
	\mathcal{C}(X_m, X_n, \omega_k) = \frac{|\hat f_{m, n} (\omega_k)|^2}{\hat
		f_{m, m}(\omega_k) \hat f_{n, n} (\omega_k)}
\end{equation}

\noindent This yields a feature matrix $\mathcal{C} \in \R^{7 \times 7 \times 129}$,
representing the functional connectivity of EEG channels at each frequency bin.
We use the vector $\mathcal{C}_{m, n}$ to represent the edge weight between any
two channels in the $7 \times 7$ fully connected graph represented by
$\mathcal{C}$, and use the STFT values $\mathcal{F}_m \in \R^{129 \times 62}$
calculated for each of our $m \in \{1, \cdots, 7\}$ channels as in
Section~\ref{subsec:stft} as the node features for our graph representation.

\subsubsection{Topological Data Analysis Features}

Topological features were extracted using persistent homology applied to EEG
coherence matrices as described in Subsection~\ref{subsec:coherence}. Recall
that based on our coherence matrix $\mathcal{C}$, two perfectly correlated
signals have coherence 1, while two completely uncorrelated signals have
coherence 0. In order to create a coherence dissimilarity matrix, we calculate a new
matrix $\mathcal{D} = 1 - \mathcal{C}$, yielding a matrix where two perfectly
correlated signals have dissimilarity 0, while two perfectly uncorrelated signals
have dissimilarity 1. We then apply \emph{persistent homology} to our coherence
dissimilarity matrix, using a Rips filtration to generate a persistence diagram for
each frequency band. For further details, we follow the approach
of~\cite{sathyanarayanaTopologicalDataAnalysis2025a}.

Persistent homology aims to measure the topological invariants of a space, as
measured through homology groups. Homology measures topological invariants such
as connected components (0th homology group or $H_0$), holes (1st homology group
or $H_1$), and other properties in higher dimensions. For this work, we focus on
$H_0$ and $H_1$. We first build a \emph{filtration} of our functional
connectivity, formally a Rips filtration. A filtration is a sequence of nested
subsets of data. We form these subsets by varying a filtration parameter, and
considering any two points with distance as defined by $\mathcal{D}$ less than
the filtration parameter as connected. From the changing connectivity pattern as
we increase the filtration parameter, we can calculate the topological features
of the data. The scalar filtration parameter value where a particular topological
feature is first observed is called the birth value, and the scalar filtration
parameter value where the topological feature is no longer observed is called
the death value. These (birth, death) pairs can be plotted in $\R^2$ to form an
object called a persistence diagram which represents the topological features of
the coherence dissimilarity space. The persistence diagram is a multiset of all
birth and death values observed from a space. Further information on the persistent
homology pipeline can be found in~\cite{edelsbrunnerComputationalTopologyIntroduction2010}.

Persistence diagrams are poorly suited to machine learning methods, as they
cannot be embedded in a Hilbert space as required by many machine learning
algorithms~\cite{bubenikStatisticalTopologicalData2015}. Recent research has
proposed Hermite Expansion of Persistence Curves (HEPC) and Arbitrary Period
Fourier Approximation of Persistence Curves (AP-FAPC) to featurize persistence
diagrams appropriately for machine learning. In order to construct these
representations, we must first construct a persistence curve.

\begin{definition}[Lifespan Entropy Persistence Curve~\cite{chungTopologicalDataAnalysis2024a}]
	Define the function $L: \mathbb{D} \rightarrow \R$, with $\mathbb{D}$ the
	space of persistence diagrams.
	\begin{equation}\label{eq:L_func}
		L = \sum_{(b_i, d_i) \in D} (d_i - b_i)
	\end{equation}
	for a persistence diagram $D$. We define our $\psi(b,d)$ function using Equation~\eqref{eq:L_func}:
	\begin{equation}
		\psi(b, d) = -\frac{d-b}{L}\log\left(\frac{d-b}{L}\right)
	\end{equation}
	We define the Lifespan Entropy Persistence Curve as:
	\begin{equation}\label{eq:lifespan_entropy}
		P(D)(x) = -\sum_{(b_i, d_i) \in D} \frac{d_i - b_i}{L}\log\left(\frac{d_i - b_i}{L}\right) \chi_{[b_i, d_i)}(x)
	\end{equation}
	for $x \in \R$.
\end{definition}

The persistence curve is a compactly supported step function on $\R$. Once we
have this function which can approximate a persistence diagram appropriately for
machine learning applications, we are left to decide how to vectorize it. Two
popular approaches exist: Hermite Expansion of Persistence Curves (HEPC), and
Arbitrary Period Fourier Approximation of Persistence Curves
(AP-FAPC)~\cite{chungStableTopologicalFeature2022a,manjunathPediatricSleepStaging2025a}.
Briefly, these approximate the persistence curve using Hermite function
coefficients in the case of HEPC, and Fourier coefficients in the case of
AP-FAPC. Prior work has shown AP-FAPC to complement HEPC approximations and lead
to greater accuracy when combined together, as AP-FAPC allows for finer grain
approximation of the underlying persistence curve. We calculate 16
coefficients for each featurization type. In the case of HEPC, we obtain an
array of size $\mathcal{F} \in \R^{2 \times 129 \times 16}$, for two homology
groups $H_0$ and $H_1$, 129 frequency bins, and 16 coefficients. Each AP-FAPC
coefficient was decomposed into real and imaginary components, yielding 16
additional coefficients per homology dimension per frequency bin. The final
feature vector concatenated HEPC coefficients with real and imaginary parts of
AP-FAPC (16 + 16 + 16 = 48 coefficients per homology group per frequency bin),
resulting in a feature array of dimensions $\mathcal{F} \in \R^{2 \times 129
		\times 48}$ that approximates the persistence curve shape.

\subsection{Model Architectures}

\subsubsection{Vision Transformer}

As a primary architecture for all features aside from Coherence-based graph
features, we use a Vision Transformer adapted for multichannel EEG
analysis~\cite{dosovitskiyImageWorth16x162020}. The model begins with instance
normalization across channels, followed by patch extraction that divides the
input into separate patches. For 2D inputs (STFT spectrograms, TDA features),
patches of size $16 \times 16$ were extracted from inputs of shape $(C, H, W)$,
while for 1D inputs (raw signals), patches of size 256 were extracted from
inputs of shape $(C, L)$. Each patch was linearly projected to a
1024-dimensional embedding space and combined with learnable positional
embeddings to preserve spatial information. The core architecture consisted of
24 transformer blocks, each containing layer normalization, multihead
self-attention with 16 attention heads, followed by a two layer MLP. We use
global average pooling to aggregate information across all patches, followed by a
two layer MLP head and a final linear classifier for binary classification. For
HEPC topological features with shape $(2, 129, 16)$ representing 2 homology
dimensions $\times$ 129 frequency bins $\times$ 16 coefficients, the model treated the two
homology groups ($H_0$ and $H_1$) as channels, extracting patches from the
frequency and coefficient representation to learn relationships between
topological features across frequency bands.

\subsubsection{Graph Neural Network}

In order to optimally process our coherence-based graph features, we use a Graph
Attention Network (GAT)~\cite{brodyHowAttentiveAre2021}. Edge features were
processed through a fully connected encoder network, while node features were
initially projected to 512 dimensions. The core architecture consisted of four
stacked GATv2 layers with eight attention heads and hidden dimensions of 512. We
use multihead attention mechanisms to weight information from neighboring
channels based on both node features and coherence edge attributes. We use
pooling with learnable attention to aggregate node representations, followed by
a fully connected classifier for binary classification. This architecture
enabled the model to capture both local channel interactions and global brain
connectivity patterns relevant to sleep apnea detection.

\subsection{Model Training}

We use the Adam optimizer to train our
models~\cite{kingmaAdamMethodStochastic2017d}. In order to combat the
significant class imbalance (~6\% apnea label vs ~94\% normal breathing label)
we use inverse frequency class weighting on our loss function. In order to
characterize the performance of our models, we used balanced accuracy (BA) and
area under receiver operating characteristic curve (AUC) as evaluation metrics.
These two metrics allow us to characterize the performance of our model on both
the majority and minority classes, and have been extensively used in biomedical
machine learning literature~\cite{tholkeClassImbalanceShould2023a}.
Hyperparameter tuning was performed through grid search. All models were trained
for 10 epochs using a one-cycle learning
rate~\cite{smithCyclicalLearningRates2017}. All models were trained with batch
size 256, except for the GAT model, which used a batch size of 1024. The models
trained in STFT and EEG timeseries data used a learning rate of 0.00001, while
the models trained on HEPC and HEPC + AP-FAPC features were both trained using a
learning rate of 0.0001. The GAT model was trained with a learning rate of
0.000001. These hyperparameters were selected by grid search.

To ensure that the model receives appropriate training data from each
demographic of subjects, we use a subject-wise stratified train/test split,
with 75\% train/25\% test split. This ensures that an equal proportion of the
training and testing data is represented by each age and sex group, i.e. the
same ratio of age 5 male subjects is included in the training set and the
testing set. Furthermore, the data split is conditioned on subjects, meaning
that all data from each subject is in either the train or test sets, but not
both. For this manuscript, we evaluate on a single train/test split; K-fold
cross validation is left for later work.

\section{Results and Discussion}

\subsection{Overall Comparative Performance Analysis}

\begin{table*}[htbp]
	\caption{Overall Performance Comparison Across Methods and Feature Representations}
	\label{tab:overall_performance}
	\centering
	\begin{tabular}{llccccc}
		\toprule
		Model       & Feature         & \# Params   & AUC            & Bal. Acc       & Sens           & Spec           \\
		\midrule
		Transformer & EEG Timeseries  & 308,376,577 & 0.639          & 0.606          & 0.609          & 0.603          \\
		Transformer & STFT            & 308,370,433 & 0.736          & 0.670          & 0.752          & 0.588          \\
		GAT         & Coherence Graph & 9,979,129   & 0.748          & \textbf{0.688} & \textbf{0.767} & 0.610          \\
		Transformer & HEPC            & 307,043,329 & \textbf{0.750} & 0.685          & 0.732          & \textbf{0.639} \\
		Transformer & HEPC + AP-FAPC  & 307,059,713 & 0.740          & 0.674          & 0.728          & 0.619          \\
		\bottomrule
	\end{tabular}
\end{table*}

Table~\ref{tab:overall_performance} presents overall test set performance. We
report area under ROC curve (AUC), balanced accuracy (Bal. Acc.), sensitivity
(Sens), and specificity (Spec), as well as the number of parameters in the model
(\# Params).
Topological features (HEPC) achieved the highest performance (AUC 0.750),
narrowly outperforming the GAT model trained on coherence representations and
exceeding traditional time-frequency features. The GAT demonstrated the
highest sensitivity and balanced accuracy despite using 96.8\%
fewer parameters than transformer-based approaches. This demonstrates that
explicit encoding of EEG channel relationships as graph structure is
substantially more parameter-efficient than learning these relationships
implicitly through transformer self-attention.
Notably, augmenting HEPC with AP-FAPC coefficients degraded performance (AUC
0.740), suggesting the additional topological complexity introduces noise rather
than complementary information. The raw EEG baseline continued to show poor
performance (AUC 0.639), confirming the necessity of structured feature
representations for this task.

The superior performance of HEPC features and the GAT model suggests that graph
representations of EEG coherence patterns capture sleep apnea signatures more
effectively than direct spectral analysis. Recall that the HEPC features
capture the topological features of our coherence graph, without including the
extra STFT node features passed to the GAT. Persistent homology extracts
multiscale topological features that may better characterize the disrupted
functional connectivity patterns associated with apnea events and cortical
arousals. The Hermite expansion provides a parameterization of persistence
curves that appears well suited to the coherence dissimilarity matrices
computed from multichannel EEG. While a best-performance AUC of 0.750 is not
sufficient for clinical application, this study represents one of the first
analyses of multiple features across similar architectures for the task of
sleep apnea identification from EEG. Further work is required to improve
classification performance.

\subsection{Performance Across Patient Demographics}

\begin{figure*}[htbp]
	\centering
	\includegraphics[width=\textwidth]{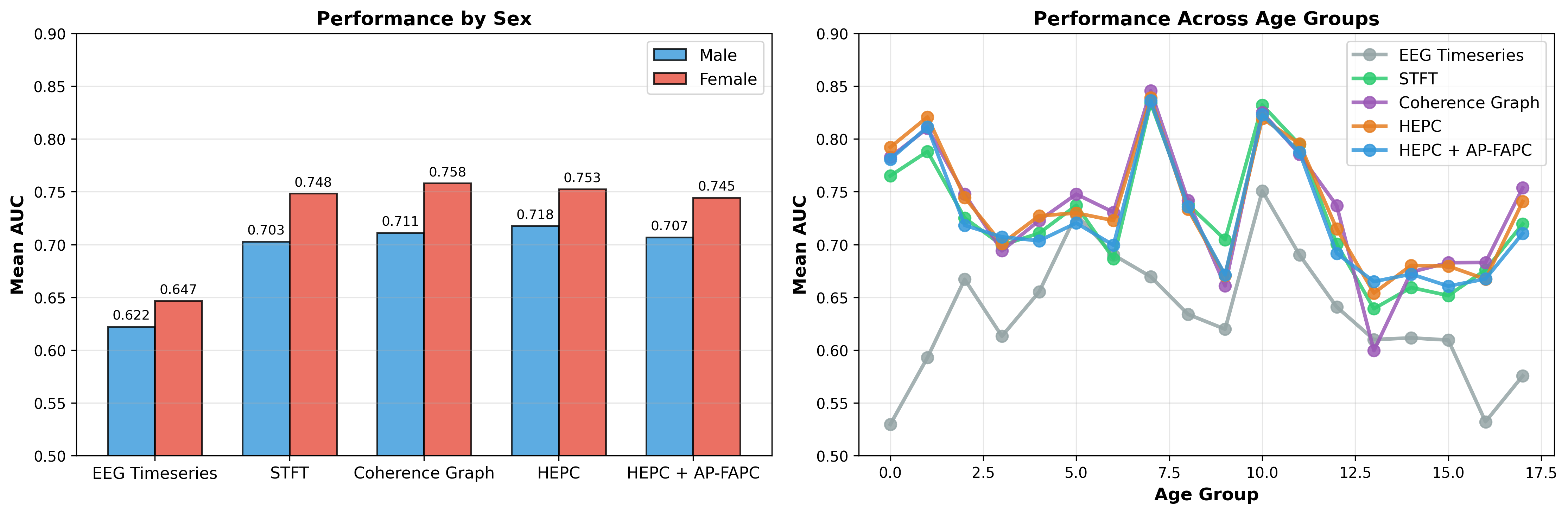}
	\caption{Model performance across age and sex groups}
	\label{fig:demographics}
\end{figure*}

Figure~\ref{fig:demographics} presents age and sex stratified model performance.
Female subjects demonstrated consistently superior performance across all
successful models. This sex-based disparity is present in all model types. The
raw EEG baseline exhibited a gap between the male and female groups alongside
poorest overall performance, indicating that the sex bias is not an artifact of
specific feature representations but rather reflects underlying differences in
either OSA presentation or EEG signal characteristics between male and female
subjects. This result is even more surprising given that we had significant
numbers of both male and female subjects (330 male, 245 female), similar AHI
levels in each group (average AHI 2.211 for males and 2.117 for females), and no
statistically significant difference in age distributions ($p = 0.416$ for a
Mann-Whitney U-test of male and female age distribution in the test set).

Performance on individual age groups peaked at age 8 and 10, while age groups
2-6 and 13-18 showed notably degraded performance. All models exhibited
performance fluctuations across age groups. The raw EEG model showed the most
extreme fluctuation. These findings underscore the importance of demographic
aware model validation, as aggregate performance metrics obscure substantial
heterogeneity that could impact clinical deployment across diverse patient
populations.

\subsection{Performance Across Sleep Apnea Severity}\label{subsec:severity_results}

\begin{table}[htbp]
	\caption{Model Performance Stratified by OSA Severity}
	\label{tab:ahi_performance}
	\centering
	\scriptsize
	\begin{tabular}{llccccc}
		\toprule
		Feature         & Severity & n   & AUC            & Bal. Acc       & Sens           & Spec           \\
		\midrule
		EEG Timeseries  & Healthy  & 383 & 0.650          & 0.615          & 0.629          & 0.602          \\
		                & Mild     & 133 & 0.651          & 0.615          & 0.616          & \textbf{0.614} \\
		                & Moderate & 33  & 0.592          & 0.564          & 0.543          & \textbf{0.585} \\
		                & Severe   & 26  & 0.590          & 0.565          & 0.558          & 0.571          \\
		\midrule
		STFT            & Healthy  & 383 & 0.748          & 0.680          & 0.758          & 0.601          \\
		                & Mild     & 133 & 0.731          & 0.671          & 0.780          & 0.562          \\
		                & Moderate & 33  & 0.664          & 0.605          & 0.709          & 0.500          \\
		                & Severe   & 26  & 0.706          & \textbf{0.650} & \textbf{0.694} & 0.606          \\
		\midrule
		Coherence Graph & Healthy  & 383 & \textbf{0.759} & \textbf{0.704} & \textbf{0.781} & 0.626          \\
		                & Mild     & 133 & \textbf{0.754} & \textbf{0.697} & \textbf{0.819} & 0.575          \\
		                & Moderate & 33  & 0.670          & 0.623          & 0.722          & 0.524          \\
		                & Severe   & 26  & 0.697          & 0.623          & 0.597          & \textbf{0.649} \\
		\midrule
		HEPC            & Healthy  & 383 & 0.754          & 0.691          & 0.725          & \textbf{0.657} \\
		                & Mild     & 133 & 0.743          & 0.681          & 0.749          & \textbf{0.614} \\
		                & Moderate & 33  & \textbf{0.706} & \textbf{0.645} & \textbf{0.780} & 0.510          \\
		                & Severe   & 26  & \textbf{0.707} & 0.641          & 0.660          & 0.623          \\
		\midrule
		HEPC + AP-FAPC  & Healthy  & 383 & 0.744          & 0.679          & 0.723          & 0.635          \\
		                & Mild     & 133 & 0.743          & 0.682          & 0.759          & 0.605          \\
		                & Moderate & 33  & 0.689          & 0.613          & 0.750          & 0.475          \\
		                & Severe   & 26  & 0.690          & 0.625          & 0.641          & 0.609          \\
		\bottomrule
	\end{tabular}
\end{table}

OSA severity stratification presented in Table~\ref{tab:ahi_performance}
revealed distinct performance patterns across the disease spectrum. Bold values
indicate best performance for an AHI severity level. Across STFT,
coherence graph, HEPC, and HEPC+AP-FAPC models, performance was highest on
healthy subjects and progressively degraded with mild and moderate OSA. However,
for the STFT and coherence graph models, the severe OSA increased significantly
from the healthy and mild groups. For the HEPC and HEPC+AP-FAPC models, the
model performance on the severe AHI remained similar to the performance in the
moderate AHI level.

\subsection{Performance Across Sleep Stages}

\begin{table}[htbp]
	\caption{Model Performance Stratified by Sleep Stage}
	\label{tab:sleep_stage_performance}
	\centering
	\scriptsize
	\begin{tabular}{lccccc}
		\toprule
		Feature         & Stage & AUC            & Bal. Acc       & Sens           & Spec           \\
		\midrule
		EEG Timeseries  & Wake  & 0.545          & 0.539          & 0.619          & \textbf{0.458} \\
		                & N1    & 0.568          & 0.552          & 0.710          & 0.395          \\
		                & N2    & 0.594          & 0.570          & 0.571          & 0.568          \\
		                & N3    & 0.539          & 0.505          & 0.090          & \textbf{0.919} \\
		                & REM   & 0.596          & 0.574          & 0.779          & 0.369          \\
		\midrule
		STFT            & Wake  & 0.688          & 0.644          & 0.857          & 0.431          \\
		                & N1    & 0.663          & 0.633          & 0.803          & 0.462          \\
		                & N2    & 0.704          & 0.636          & 0.684          & 0.589          \\
		                & N3    & 0.705          & 0.614          & 0.491          & 0.737          \\
		                & REM   & 0.705          & 0.654          & 0.766          & 0.541          \\
		\midrule
		Coherence Graph & Wake  & \textbf{0.702} & \textbf{0.649} & \textbf{0.900} & 0.398          \\
		                & N1    & 0.664          & \textbf{0.642} & \textbf{0.845} & 0.439          \\
		                & N2    & 0.697          & 0.641          & \textbf{0.718} & 0.564          \\
		                & N3    & 0.697          & 0.564          & 0.242          & 0.886          \\
		                & REM   & 0.712          & \textbf{0.668} & \textbf{0.798} & 0.538          \\
		\midrule
		HEPC            & Wake  & 0.699          & 0.648          & 0.868          & 0.428          \\
		                & N1    & \textbf{0.678} & \textbf{0.642} & 0.789          & \textbf{0.495} \\
		                & N2    & \textbf{0.719} & \textbf{0.646} & 0.623          & \textbf{0.669} \\
		                & N3    & 0.732          & 0.637          & 0.478          & 0.795          \\
		                & REM   & \textbf{0.721} & 0.666          & 0.754          & 0.578          \\
		\midrule
		HEPC + AP-FAPC  & Wake  & 0.693          & 0.645          & 0.857          & 0.434          \\
		                & N1    & 0.670          & 0.635          & 0.781          & 0.489          \\
		                & N2    & 0.708          & 0.634          & 0.629          & 0.638          \\
		                & N3    & \textbf{0.734} & \textbf{0.647} & \textbf{0.538} & 0.757          \\
		                & REM   & 0.715          & 0.655          & 0.726          & \textbf{0.583} \\
		\bottomrule
	\end{tabular}
\end{table}

Sleep stage stratification presented in Table~\ref{tab:sleep_stage_performance}
revealed performance variations across sleep stages. Bold values indicate best
performance across sleep stages. N2 and N3 sleep stages demonstrated the highest
overall performance, with HEPC achieving AUC in N2, and HEPC+AP-FAPC achieving
highest AUC in N3 sleep. Wake and N1 light sleep showed poorest aggregate
performance. Notably, wake and N1 epochs produced high sensitivity but poor
specificity , indicating systematic over prediction of disordered breathing
during arousal-prone states. Conversely, N3 deep sleep yielded lower sensitivity
but high specificity, suggesting models miss apnea events that fail to trigger
cortical arousals. This sensitivity and specificity profile across the sleep
stages demonstrates that EEG-based detection fundamentally captures
arousal-associated apnea rather than pure respiratory obstruction.

The sleep stage stratification results further suggest that our models
primarily detect EEG signatures associated with arousal-triggering apnea events
rather than respiratory obstruction. Across all successful models, we observe
an inverse relationship between sensitivity and specificity: Wake and N1 stages
exhibit high sensitivity ($>0.80$ for STFT, coherence graph, and HEPC) but poor
specificity ($<0.50$), while N3 stage shows the opposite pattern with sensitivity
dropping to 0.478 (HEPC) or lower, while specificity rises to 0.795-0.886. This
trade-off aligns with known physiology of arousal thresholds across sleep
stages. In light sleep, cortical arousal is easily triggered, the model flags
many true positive events but also misclassifies non-apnea arousals as
disordered breathing~\cite{eckertArousalSleepImplications2014}. Conversely, in
N3 sleep, where the arousal threshold is physiologically higher and many apnea
events fail to produce a cortical response, the model misses most events yet
maintains high specificity when it does detect
something~\cite{eckertArousalSleepImplications2014}. Our positive labels are
defined by technician-annotated apnea/hypopnea and oxygen desaturation events
rather than isolated arousals. The classifiers appear to have learned a
combination of cortical activation patterns and disrupted connectivity that
accompanies respiratory events severe enough to trigger arousal. This
interpretation is supported by the superior performance of TDA features (HEPC),
which capture global network topology of coherence rather than local arousal
transients, suggesting the model exploits sustained, multiscale changes in
functional connectivity induced by intermittent hypoxia and sleep
fragmentation.

\section{Conclusion}

In this work, we present a comprehensive evaluation of deep learning
architectures and feature representations for automated sleep apnea detection
from multichannel EEG, with rigorous stratification across patient
demographics, OSA severity levels, and sleep stages. Topological data analysis
features (HEPC) achieved the highest overall performance,
outperforming traditional time-frequency representations and
demonstrating that persistent homology effectively captures central nervous
system manifestations of sleep-disordered breathing. However, demographic
stratification revealed critical performance heterogeneity: models consistently
performed better on female subjects, model performance on moderate OSA was
unexpectedly poor, and sleep stage analysis uncovered high sensitivity and low
specificity in lighter sleep and low sensitivity and high specificity in deeper
sleep. These findings demonstrate that EEG-based detection fundamentally capture
arousal-associated apnea rather than respiratory obstruction, with performance
strongly dependent on sleep architecture and patient characteristics. While our
results advance automated OSA detection methodology and establish topological
features as a promising approach, the substantial performance variability across
demographic and physiological subgroups underscores the need for demographic
aware model development and larger multicenter validation before clinical
deployment. Future work should validate these approaches on external datasets
and explore ensemble methods leveraging complementary features across patient
populations.

\bibliographystyle{plain}
\bibliography{references}

\end{document}